# User Dependent Features in Online Signature Verification

**D. S. Guru, K. S. Manjunatha and S. Manjunath**

**Abstract** In this paper, we propose a novel approach for verification of on-line signatures based on user dependent feature selection and symbolic representation. Unlike other signature verification methods, which work with same features for all users, the proposed approach introduces the concept of user dependent features. It exploits the typicality of each and every user to select different features for different users. Initially all possible features are extracted for all users and a method of feature selection is employed for selecting user dependent features. The selected features are clustered using Fuzzy C means algorithm. In order to preserve the intra-class variation within each user, we recommend to represent each cluster in the form of an interval valued symbolic feature vector. A method of signature verification based on the proposed cluster based symbolic representation is also presented. Extensive experimentations are conducted on MCYT-100 User (DB1) and MCYT-330 User (DB2) online signature data sets to demonstrate the effectiveness of the proposed novel approach.

D. S. Guru (✉) · K. S. Manjunatha
Department of Studies in Computer Science, Manasagangothri, University of Mysore, Mysore 570006, Karnataka, India
e-mail: dsg@compsci.uni-mysore.ac.in

K. S. Manjunatha
e-mail: kowshik.manjunath@gmail.com

S. Manjunath
Postgraduate Department of Studies in Computer Science, JSS College of Arts, Commerce and Science, Mysore 570025, Karnataka, India
e-mail: manju_uom@yahoo.co.in

# 1 Introduction

Signature is the most widely accepted behavioral biometric trait for personal and document authentication in many day-to-day applications. Signature verification has been an active area of research due to its wide acceptance for authentication purpose in many financial and legal transactions like validation of bank cheques, credit card transactions, contracts, and bonds. Depending on the acquisition method, signatures are of two categories offline and online [1]. An offline signature is the static image of the signature available in documents and captured using devices like camera and scanner to have its digital representation. It is easy to forge an off-line signature. In an off-line signature only X and Y co-ordinates (static features) of the signature image are available for verification. On the other hand, online signatures are captured using special devices like digitizing tablet and smart pens which captures both dynamic properties and shape information. Due to the availability of additional dynamic features such as pressure, azimuth, speed, total signature time etc., it is difficult for the forger to imitate both the image shape and the way it has been originally written by the authentic signer and hence it is more reliable compared to offline signature. In an online mode the signature is represented as a time function of various dynamic properties like pressure v/s time [2].

The two stages in any biometric system are (a) Identification (b) Verification [3]. In identification, the test signature is compared with the reference signatures of all the N users in the knowledgebase to establish the identity of the signer. It is a 1: N matching process and hence takes a longer time. In verification, the test signature of the claimed user is compared with other signatures of the claimed user available in the knowledgebase to determine if the given signature is genuine.

Online signature verification is broadly classified into two categories—Parametric approach and Function based approach [4]. In the parametric approach, set of parameters obtained from the signatures are used as the representative of the particular signature. Position, speed, acceleration, number of pen ups and pendowns, pen down time ratios are the parameters proposed in the literature for online signature verification [5, 6]. In function-based approach, signature is represented as time function of various dynamic properties. In the first category, during verification, the parameters of query and reference signatures are compared to determine whether the query signature is genuine or not. In the function-based approach, query and reference signatures are compared either point-to-point or segment-to-segment basis [7]. A function based approach takes more time as it involves comparing every point in the signature trajectory but gives better performance. In the work of [8], parametric approach shows equally competitive result compared to any function based approach.

Features for on online signatures are categorized into two types (a) local features, which are extracted from specific point in the signature (b) global features describe the whole signature or major part of the signature. Some of the local features for online signatures are curvature change, pressure, speed etc., while the global features are signature writing time, number of strokes, average speed etc., [1].

To establish the authenticity of a test signature, during matching a test signature is compared with the reference signatures stored in the knowledgebase. Different matching techniques proposed in the literature for online signatures are Hidden Markvov Model [9], Support Vector Machine [10], Neural Network [11] and Dynamic Time warping [1] and symbolic classifier [12, 13].

Almost all the signature verification methods proposed in the literature have utilized same features either local or global features for all users. However, signature is a complex biometric trait where each user has his/her own style of signing and hence the same features may not be effective in capturing the typicality of individual user. To the best of our knowledge and from literature survey the concept of user dependent features is not utilized for signature verification. In addition, signature samples of a class have large intra-class variation and there is a need to capture this intra-class variation using suitable representation scheme.

Few works are reported in literature for effective capturing of intra-class variation in signatures. Guru et al. [13] used the concept of cluster based symbolic representation for signature representation which effectively captures intra-class variation. But they have used all the 100 features for all the users which is computationally expensive. In this paper, we propose user dependent features for online signatures. Initially all possible features are extracted for all users and a method of feature selection is employed for selecting user dependent features followed by clustering of signatures based on the features selected. Clustering provides an effective representation in the form of multiple reference signatures for each class. A method of signature verification based on the proposed representation is presented. Instead of storing every signature sample of every user in the database, training signatures are clustered into a number of clusters and each cluster is stored in the knowledgebase by means of symbolic feature vector. The major contribution of this work relies on proposal of user dependent features for signatures which vary from a user to a user instead of a set of common features for all users.

The paper is organized as follows: The proposed model is explained in Sect. 2. In Sect. 3, we summarize the details of experimentation along with the result obtained. A comparative study of our work with other similar work is presented in Sect. 4. Finally in Sect. 5 some conclusions are drawn.

## 2 Proposed Model

The proposed model has three stages, user dependent feature selection, cluster based symbolic representation and signature verification based on the proposed symbolic representation.

### 2.1 User Dependent Feature Selection

Traditional feature selection methods are either supervised or unsupervised. In supervised mode, features are selected such that the importance of each feature is evaluated by the correlation between class labels and features. Some of the supervised feature selection methods include Pearson correlation coefficient, Fisher score, and information gain. In an unsupervised feature selection method, top ranked features are selected based on a certain score computed for each feature. Here correlation between feature and class labels is neglected and hence the feature selected may not be optimal. In this section we exploit an unsupervised feature selection method suitable for multi-cluster data [14]. Features selected preserve the multi-cluster structure. Tradition feature selection problem selects the features based on certain evaluation criteria, which is computationally expensive as it is a combinatorial optimization problem. The feature selection method we adapted is computationally efficient as it involves a sparse Eigen-problem and L1-regularized least square problem. It uses spectral analysis technique to measure the correlation between different features without class label information.

In spectral clustering, data points are clustered using top eigenvectors of graph laplacian, which is defined on the affinity matrix of the data points. From the perspective of graph partitioning it finds the best cut of the graph so that the criterion function can be optimized. Spectral clustering basically consists of two steps. The first step is "unfolding" the data manifold using the manifold learning algorithm and the second step performing traditional clustering on the "flat" embedding for the data points. The different steps in the feature selection algorithm that we adapted in our work are:

1. Initially graph with one vertex for each data point is created. For each data point $x_i$, p nearest neighbors are identified and an edge is drawn between each data point and all its neighbors. A weight matrix (W) based on the weight of each edge is created. In the weight graph, one of the three weighting scheme can be used: 0–1 weighting, Heat-kernel weighting, dot product weighting [14].
2. From the weight matrix ($W$), a diagonal matrix $D$ is computed whose entries are column or row sums of $W$. $D_{ii} = \sum W_{ij}$.
3. Corresponding graph Laplacian is obtained as $L = D - W$.
4. The "flat"embedding for the data points which "unfold" the data manifold can be found by solving the following generalized eigen-problem $Ly = \lambda Dy$.

Let $Y = [y_1, \ldots y_k]$, $y_i$'s are the eigenvectors of the above generalized eigenproblem with respect to the smallest eigenvalue. Solving the corresponding eigenproblem of step 4 results in a set of eigen vectors corresponding to smallest eigen values. $K$ indicates the dimensionality of the data and each $y_i$ reflects the distribution of data on the corresponding cluster.

After obtaining flat embedding $Y$ for data points, the importance of each feature for differentiating each cluster is measured. Given $y_i$, a column of $Y$, we can find a relevant subset of features by minimizing the fitting error as follows:

$$\min_{a_i} \left\| y_i - X^T a_i \right\|^2 + \beta |a_i| \qquad (1)$$

Each $a_i$ contains the combination coefficients for different features in approximating $y_i \cdot |a_i|$ is the $L - 1$ norm of $a_i$ and $X$ is the set of data points.

The advantage of using a $L - 1$ regularized regression model is to find the subset of features instead of evaluating the contribution of each feature. Each $a_i$ essentially contains the combination of coefficients for different features. It helps in approximating a subset containing the most relevant features corresponding to the non-zero coefficients in $a_i$ with respect to $y_i$. Equation (1) is essentially a regression problem and can be solved using Least Angle regression (LAR) algorithm which results in $K$ sparse coefficient vector $a_i$. Each entry in $a_i$ is a feature and the cardinality of $a_i$ is $d$ which denotes the number of features to be selected.

From the sparse coefficient vector, $d$ features are selected by computing MCFS score with respect to each feature and top $d$ features are selected in the decreasing order of MCFS score.

In our work, we exploited the feature selection method described above for selecting different features for different users. In our proposed method, signature data set of dimension $N \times M \times K$ where $N$ is the number of users, $M$ is the number of samples and $K$ is the number of features is decomposed into $N$ feature vectors of size $M \times K$. Feature selection method discussed above is applied on each of these feature vectors representing an individual user separately. It results in the reduction of dimension of feature vector of a user to a size $M \times d$ where $d$ is the number of features selected $(d < k)$. The $d$ number of features selected varies from a user to user. The indices of the corresponding features selected is also stored in the knowledgebase which is used during verification. The computational complexity of user dependent feature selection is $O(N^2M + Kd^3 + NKd^2 + M \log M)$ where N is the number of samples, M is the original number of features, K is the number of clusters, d is the number of features selected. For more details on multi-cluster feature selection, readers are referred to Cai et al. [14]. Once the user dependent features are selected, we effectively capture the intra-class variation through the concept of symbolic representation [13] which is described in next section.

## 2.2 Cluster Based Symbolic Representation

Once the user dependent features are selected, training signatures of each user are clustered using the selected features instead of all the original features. Clustering is effective as it provides multiple reference signatures for each user. We have adapted Fuzzy C means for clustering [15]. After the signatures are clustered, each cluster is represented in the form of interval-valued feature vector [13]. This representation is very effective in capturing intra-class variation which is common in signatures.

Let $\{S_1, S_2, \ldots, S_n\}$ be $n$ signature samples of a cluster $C_j$, $j = 1, 2, \ldots, C$ where $C$ is the number of clusters in each class. Let $\{f_{j1}, f_{j2}, \ldots, f_{jd}\}$ be the feature vector representing the cluster $C_j$ where $d$ is the number of features. Let $M_{jk}$, $k = 1, 2, \ldots, d$ and $\sigma_{jk}$, $k = 1, 2, \ldots d$ be the mean and standard deviation of $k$th feature of the cluster $C_j$ i.e.

$$M_{jk} = \frac{1}{n}\sum_{i=1}^{n} f_{ik} \text{ and } \sigma_{jk} = \left[\frac{1}{n}\sum_{1}^{n}\left(f_{ik} - \mu_{jk}\right)^2\right]^{\frac{1}{2}} \quad (2)$$

In order to capture intra-class variation, each feature of the cluster $C_j$ is represented in the form of interval-valued feature. For example $k$th feature of the cluster $C_j$ is represented as $\left[f_{jk}^-, f_{jk}^+\right]$ where $f_{jk}^- = M_{jk} - \alpha\sigma_{jk}$ and $f_{jk}^+ = M_{jk} + \alpha\sigma_{jk}$ for some scalar $\alpha$ which is used to constrain the upper and lower limits for $k$th feature of cluster $C_j$. Thus, the interval $\left[f_{jk}^-, f_{jk}^+\right]$ depends on the mean and the standard deviation of the respective individual feature of a cluster. The interval $\left[f_{jk}^-, f_{jk}^+\right]$ represents the lower and upper limits of the $k$th feature value of a signature cluster in the knowledgebase. In general each of the $d$ features selected is represented in the form of an interval-valued feature. The reference signature for the cluster $C_j$ is thus formed as $RFC_J = \left\{\left[f_{j1}^-, f_{j1}^+\right], \left[f_{j2}^-, f_{j2}^+\right]\cdots\left[f_{jd}^-, f_{jd}^+\right]\right\}$, $j = 1, 2, \ldots C$ where $C$ is the number of clusters in each signature class.

This symbolic feature vector is stored in database as the representative of the entire cluster. Instead of storing every signature of every cluster, it is sufficient to store one feature vector for each of the cluster. If there are $C$ clusters formed for each individual user and $N$ is the number of users then we have totally $NC$ number of reference signatures in the knowledgebase instead of $Nn(> NC)$ number of signatures.

## 2.3 Signature Verification

During verification, we consider a test signature, which is represented in the form of k features of crisp type as $F_t = \{f_{t1}, f_{t2}, \ldots, f_{tk}\}$. Each feature of the test signature is of type crisp in contrast with a reference signature where the corresponding feature is of type interval valued. For authenticating the test signature, we compare the only $d$ features of a test signature with the corresponding d interval valued features of the reference signature stored in the knowledgebase. The indices of the $d$ features of test signatures to be compared with corresponding features reference signature is available in the knowledgebase. Reemploying of feature selection is not required for a test signature as the features selected for the claimed user is known at the time of training.

The total number of features of the test signature which lie within the corresponding interval valued features of a reference signature is called degree of authenticity [13]. Degree of authenticity is expressed by means of an acceptance count, which is a measure of authenticity for the test signature to qualify as genuine or forgery. If a feature of the test signature lies within corresponding interval-valued feature of reference signature, the acceptance count is incremented by one. If the total acceptance count is greater than the predefined threshold, then the test signature is accepted as genuine else, it is considered as forgery.

The acceptance count is defined to be

$$A_c = \sum_{i=1}^{d} C\left(f_{ti}, \left[f_{ji}^-, f_{ji}^+\right]\right) \tag{3}$$

where

$$C\left(f_{ti}, \left[f_{ji}^-, f_{ji}^+\right]\right) = \begin{cases} 1 & if \left(f_{ti} \geq f_{ji}^- \; and \; f_{ti} \leq f_{ji}^+\right) \\ 0 & otherwise \end{cases} \tag{4}$$

Here $\{f_{t1}, f_{t2}, \ldots f_{td}\}$ defines the feature vector of the test signature consisting of values corresponding to the selected features.

Each feature of the test signature which lies within the corresponding interval valued feature of the reference signature contributes a value of 1 towards acceptance count.

## 3 Experimentation and Result

In this section, we discuss on dataset used and on the details of experiments conducted in our work along with the result obtained.

### 3.1 Experimentation

**Dataset**: We have conducted experiments on two data sets MCYT-100 (DB1) consisting of signatures of first 100 users and MCYT-330 (DB2) consisting of signatures of all the 330 users [16]. Both MCYT-100 and MCYT-330 online signature databases consisting of 25 genuine and 25 skilled forgeries for each user. We have used a set of 100 global features of online signatures for our experimentation. The details of these 100 global features of online signature can be found in the work of [17]. The purpose of using the DB1 is to set up the system with a small set of users. Once the different parameters like similarity threshold, number of features to be selected for each user keeping EER minimum are decided with the small database DB1, the experimentations are performed on the whole database. This results in reduction of computation time and avoids the risk of over training. Feature selection experiments are repeated for each user in the training phase which selects best features for the particular user. Experiments are conducted under varying number of features.

**Experimental setup**: Initially we conducted feature selection experiments on DB1 by varying the number of features from 5 to 75 in step of 5 and noted the EER in each case. Further, experiments are conducted by varying the feature numbers in step of 1 to identify the best value of the number of features for achieving a minimum EER. Similarity threshold values are also varied from 0.1 to 0.9. We conducted 20 trials and for every trial, training signatures are randomly selected and EER is noted in each trial. Finally, we considered the average EER of all the twenty trials as the final EER value. The number of features to be selected is empirically fixed so that EER is minimum. We have used DB1 as a validation dataset for fixing up the value the number of features to be selected. Once the value of the number of features to be selected is decided, we conducted experiments on DB2 using the same value of the number of features and noted down the EER. In DB2 also, we randomly selected the training signatures in each of the 20 trials and conducted verification experiments. Details of training and testing signatures in our experimentation for both DB1 and DB2 are tabulated in Tables 1 and 2 respectively. Figure 1a–d shows the variation of FAR and FRR in all the four categories with respect to DB1.

We trained the system with a small training set consisting of 5 genuine signatures (Skilled_05 and Random_05) and with a big training set consisting of 20

**Table 1** Details of the training and testing signatures with DB1

| Training/testing samples | Skilled_05 | Skilled_20 | Random_05 | Random_20 |
|---|---|---|---|---|
| Number of training signatures | 100 users × 5 genuine signatures per user = 500 | 100 users × 20 genuine signatures per user = 2,000 | 100 Users × 5 genuine signatures per user = 500 | 100 Users × 20 genuine signatures per user = 2,000 |
| Number of testing signature | 100 users × 20 genuine signatures per user = 2,000 + 100 users × 25 skilled forgery per user = 2,500 | 100 users × 5 genuine signatures per user = 500 + 100 users × 25 skilled forgery = 2,500 | 100 users × 20 Genuine signatures per user = 2,000 + 100 users × 99 random forgeries = 9,900 | 100 users × 5 Genuine signatures per user = 500 + 100 users × 99 random forgeries = 9,900 |

**Table 2** Details of the training and testing signatures with DB2

| Training/testing samples | Skilled_05 | Skilled_20 | Random_05 | Random_20 |
|---|---|---|---|---|
| Number of training signatures | 330 users × 5 genuine signatures per user = 1,650 | 330 users × 20 genuine signatures per user = 6,600 | 330 Users × 5 genuine signatures per user = 1,650 | 330 User × 20 genuine signatures per user = 6,600 |
| Number of testing signature | 330 users × 20 genuine signatures per user = 6,600 + 330 users × 25 skilled forgery per user = 8,250 | 330 users × 5 genuine signatures per user = 1,650 + 330 users × 25 skilled forgeries per user = 8,250 | 330 users × 20 genuine signatures per user = 6,600 + 330 user × 329 random forgeries = 108,570 | 330 users × 5 genuine signatures per user = 1,650 + 330 user × 329 = 108,570 |

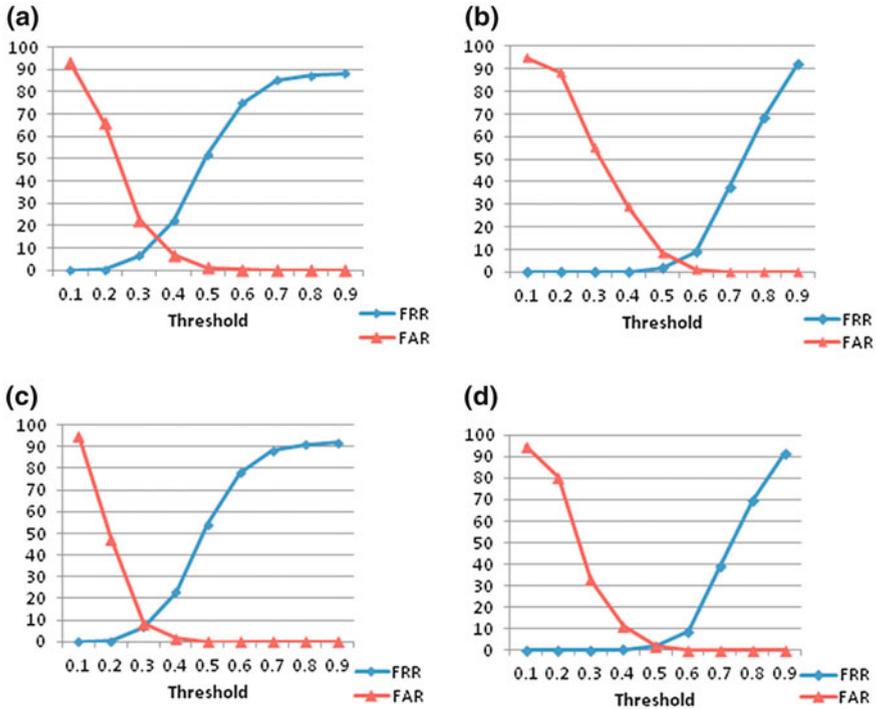

**Fig. 1** Variation of FAR and FRR under varying thresholds for DB1: **a** For Skilled_05 with 60 features. **b** For Skilled_20 with 50 features. **c** For Random_05 with 60 features. **d** For Random_20 with 50 features

genuine signatures (Skilled_20 and Random_20). The test set consists of the remaining genuine signatures and all the forgery signatures. In case of random forgery, genuine signature of every other user is considered as random forgery.

## 3.2 Experimental Results

In our experimentation, training signatures are clustered using Fuzzy C-means as it is distribution free when compared to other well known statistical techniques like KNN classifier, maximum likelihood estimate etc., and also its ability to discover cluster among data. Verification performance of our method for DB1 and DB2 are shown in Table 3. Data in Table 3 shows the minimum EER achieved. The EER value for varying number of features selected in DB1 is shown in Table 4. In case of skilled-05 and Random-05, we achieved a minimum EER for 60 features. Even if the number of features selected is increased in step of 1 up to 75, there was only marginal decrease in the EER and hence we considered only 60 features and similarly with respect to Skilled_20 and Random_20 we achieved a minimum EER for 50 features.

**Table 3** Minimum EER with Skilled and Random forgery

| Dataset | Skilled_05 (60 features) | Skilled_20 (50 features) | Random_05 (60 features) | Random_20 (50 features) |
|---|---|---|---|---|
| MCYT-100(DB1) | 14.90 | 5.06 | 7.98 | 2.02 |
| MCYT-330(DB2) | 15.90 | 6.10 | 1.90 | 1.80 |

**Table 4** Verification performance (EER) with skilled and random forgeries under varying features of DB1

| Features | Skilled_05 | Skilled_20 | Random_05 | Random_20 |
|---|---|---|---|---|
| 5 | 26.05 | 17.77 | 23.28 | 13.93 |
| 10 | 21.20 | 12.86 | 19.52 | 7.63 |
| 15 | 19.19 | 8.92 | 15.73 | 5.82 |
| 20 | 17.66 | 8.08 | 14.06 | 5.77 |
| 25 | 17.55 | 7.42 | 13.48 | 4.58 |
| 30 | 16.75 | 6.98 | 12.37 | 4.22 |
| 35 | 16.68 | 6.14 | 12.07 | 3.08 |
| 40 | 16.00 | 6.00 | 10.56 | 2.77 |
| 45 | 16.20 | 5.45 | 10.45 | 2.33 |
| 50 | 15.47 | 5.06 | 9.09 | 2.02 |
| 55 | 15.47 | 5.07 | 9.015 | 2.15 |
| 60 | 14.90 | 5.22 | 7.98 | 2.00 |
| 65 | 15.05 | 4.76 | 8.01 | 1.96 |
| 70 | 14.52 | 4.91 | 7.25 | 1.86 |
| 75 | 14.42 | 4.56 | 7.305 | 1.87 |

## 4 Comparative Analysis

In this section, we compare the verification performance of our method with that of other existing methods. Since most of the researchers have reported their work on DB1, we have considered DB1 for comparative analysis. Table 5 shows the performance of different verification systems which work on the same dataset as ours. Details of some of these classifiers is found in the work of [2, 17, 18]. From Table 5 it is clear that our method is superior when compared to methods like NND, Base Classifier, MOGD_3 and MOGD_2 in Skilled_20, Random_5, Random_20 categories and when compared to methods like Symbolic classifier, LPD, PCAD and SVD, even though the EER we obtained is slightly high we have used only 50 features (Skilled_20 and Random_20) and 60 features(Skilled_05 and Random_05) while others have used all the 100 global features.

**Table 5** Equal error rates of various online signature verification approaches on DB1

| Method | Skilled_05 | Skilled_20 | Random_05 | Random_20 |
|---|---|---|---|---|
| 1. Proposed method | 14.9 | 5.0 | 7.9 | 2.0 |
| 2. Symbolic classifier [13] | 15.4 | 4.2 | 3.6 | 1.2 |
| 3. Linear programming description (LPD) | 9.4 | 5.6 | 3.6 | 2.5 |
| 4. Principal component analysis description (PCAD) | 7.9 | 4.2 | 3.8 | 1.4 |
| 5. Support vector description (SVD) | 8.9 | 5.4 | 2.8 | 1.6 |
| 6. Nearest neighbour method description (NND) | 12.2 | 6.3 | 6.9 | 2.1 |
| 7. Random ensemble of base (RS) | 9.0 | – | 5.3 | – |
| 8. Random subspace ensemble with resampling of base (RSB) | 9.0 | – | 5.0 | – |
| 9. Base classifier (BASE) | 17.0 | – | 8.3 | – |
| 10. Parzen window classifier (PWC) | 9.7 | 5.2 | 3.4 | 1.4 |
| 11. Mixture of Gaussian description_3 (MOGD_3) | 8.9 | 7.3 | 5.4 | 4.3 |
| 12. Mixture of Gaussian description_2 (MOGD_2) | 8.1 | 7.0 | 5.4 | 4.3 |
| 13. Gaussian model description S | 7.7 | 4.4 | 5.1 | 1.5 |
| 14. Kholmatov model (KHA) | 11.3 | – | 5.8 | – |
| 15. Fusion methods | 7.6 | – | 2.3 | – |
| 16. Regularized Parzen window classifier RPWC [8] | 9.7 | – | 3.4 | – |

## 5 Conclusion

In this work, we introduced a novel concept of user dependent features for online signature verification. The proposed method is very effective in capturing the typicality of individual user. In addition, our method is computationally efficient as it works in lower dimension. We conducted extensive experiments on MCYT database and the result demonstrates the effectiveness of the proposed method. Overall, the idea of adaption of user dependent features for online signature verification is our contribution which is first of its kind in the literature.

**Acknowledgments** We thank Dr. Julian Fierrez Aguillar, Biometric Research Lab-AVTS, Spain for providing MCYT Online signature dataset. We also thank Deng Cai, Associate Professor, Zhejiang University, China for sharing his work on unsupervised feature selection for multi-cluster data.